%% file: main_apple.tex
\documentclass{applemlr}

\usepackage{amsmath}
\usepackage{enumerate}
\usepackage{algorithm}
\usepackage{algpseudocode}
\usepackage{amsfonts}
\usepackage{amsthm}
\usepackage{cleveref}
\usepackage{diagbox}
\usepackage{colortbl}
\usepackage{amssymb}
\usepackage{xspace}
\usepackage{wrapfig}
\usepackage{adjustbox}
\usepackage{tabularx}
\usepackage{booktabs}
\usepackage{mathtools}
\usepackage{tikz}
\usepackage{enumitem}
\usepackage{silence}
\usepackage{dsfont}
\usepackage{multirow}
\usepackage{makecell}
\usepackage{xfakebold}
\usepackage[cachedir=minted-cache,frozencache]{minted}

\usepackage[T1]{fontenc}
\usepackage{lmodern}

\definecolor{textgray}{HTML}{6E6E73}
\usetikzlibrary{positioning, calc}
\usetikzlibrary{decorations.pathmorphing}

\makeatletter
\patchcmd{\wrong@fontshape}{\@gobbletwo}{}{}{}
\makeatother
\numberwithin{equation}{section}
\makeatletter
\AtBeginDocument{
  \urlstyle{sf}
  
}
\makeatother

\definecolor{light}{RGB}{125, 125, 125}
\crefname{tcb@cnt@pbox}{code}{code}
\Crefname{tcb@cnt@pbox}{Code}{Code}
\crefname{assumption}{assumption}{assumption}
\Crefname{assumption}{Assumption}{Assumptions}

\newtcolorbox[auto counter]{pbox}[2][]{
  colback=white,
  title=Code~\thetcbcounter: #2,
  #1,fonttitle=\sffamily,
  fontupper=\sffamily,
  arc=2pt,
  colframe=bgcolor,
  coltitle=fgcolor,
  colbacktitle=bgcolor,
  toptitle=0.25cm,
  bottomtitle=0.125cm
}

\makeatletter
\newcommand\applefootnote[1]{%
  \begingroup
  \renewcommand\thefootnote{}%
  \renewcommand\@makefntext[1]{\noindent##1}%
  \footnote{#1}%
  \addtocounter{footnote}{-1}%
  \endgroup
}
\makeatother

\definecolor{cverbbg}{gray}{0.90}

\usepackage{xeCJK}
\usepackage{polyglossia}
\usepackage{microtype}
\usepackage{graphicx}
\usepackage{booktabs,longtable,array,colortbl}
\usepackage{etoolbox}

\setdefaultlanguage{english}
\setotherlanguages{french,telugu,russian,thai}

\def\appendixautorefname{Appendix}

\makeatletter
\patchcmd{\hyper@makecurrent}{%
    \ifx\Hy@param\Hy@chapterstring
        \let\Hy@param\Hy@chapapp
    \fi
}{%
    \iftoggle{inappendix}{%
        \@checkappendixparam{chapter}%
        \@checkappendixparam{section}%
        \@checkappendixparam{subsection}%
        \@checkappendixparam{subsubsection}%
        \@checkappendixparam{paragraph}%
        \@checkappendixparam{subparagraph}%
    }{}%
}{}{\errmessage{failed to patch}}

\newcommand*{\@checkappendixparam}[1]{%
    \def\@checkappendixparamtmp{#1}%
    \ifx\Hy@param\@checkappendixparamtmp
        \let\Hy@param\Hy@appendixstring
    \fi
}
\makeatother

\newtoggle{inappendix}
\togglefalse{inappendix}

\apptocmd{\appendix}{\toggletrue{inappendix}}{}{\errmessage{failed to patch}}

\newcommand{\numlangs}{14}

\title{mAceReason-Math: A Dataset of High-quality Multilingual Math Problems Ready for RLVR}
\author{Konstantin Dobler$^{*,\ddagger,\S}$}
\author{Simon Lehnerer$^{*,\dagger}$}
\author{\\Federico Scozzafava$^{\dagger}$}
\author{Jonathan Janke$^{\dagger}$}
\author{Mohamed Ali$^{\dagger}$}

\affiliation{$^\dagger$Apple}
\affiliation{$^\S$Hasso Plattner Institute \& ELLIS Unit Potsdam}
\affiliation{\\$^*$Equal contribution}
\affiliation{$^\ddagger$Work done during an internship at Apple}

\usepackage{catchfile}
\abstract{\input{content_abstract.tex}}

{
\metadata[Data]{
\url{https://github.com/apple/ml-macereason-math}}
\metadata[Correspondence]{\sffamily konstantin.dobler@hpi.de}
\date{\sffamily March 11, 2026}
}
\begin{document}

\maketitle

\input{content}

\section*{Limitations}
We employ an LLM-based translation and refinement pipeline, so residual translation or formatting errors may persist despite extensive quality control. We validate the pipeline with native-speaker annotators, but this excludes Swahili, Telugu, and Bengali due to a lack of available annotators. Translation quality may therefore be more variable for these languages. Moreover, only the test set received full native-speaker review of the final translations (again excluding Swahili, Telugu, and Bengali).
Our evaluation also depends on automated answer verification tools\footnote{Such as \url{https://github.com/huggingface/Math-Verify}.}, which do gracefully handle non-US number formatting in many cases (\emph{e.g.}, commas as decimal separators) but may be less reliable for complex expressions, potentially leading to occasional false negatives. Finally, to limit API usage, we evaluate closed models only at pass@1 with a single rollout per prompt.

\section*{Acknowledgments}
We thank Adam Golinski and Michael Kirchhof for their insightful discussion and valuable feedback.
We also thank our team of professional translators and native speaker validators who ensured high-quality translations across all supported languages.

\bibliographystyle{plainnat}
\bibliography{references}

\appendix
\clearpage

\input{content_appendix}

\applefootnote{ \textcolor{textgray}{\sffamily Apple and the Apple logo are trademarks of Apple Inc., registered in the U.S. and other countries and regions.}}

\end{document}

%% file: content_abstract.tex
Reinforcement Learning with Verifiable Rewards (RLVR) has been successfully applied to significantly boost the capabilities of pretrained large language models, especially in the math and logic problem domains. However, current research and available training datasets remain English-centric. While multilingual training data and benchmarks have been created in the past, they were not created with RLVR and current model capability in mind, and their level of difficulty is often too low to provide appropriate training signals for current models.
To address this gap, we provide \textbf{mAceReason-Math}, a dataset of high-quality translations of challenging math problems sourced from a corpus specifically curated for RLVR (AceReason-Math). We further take specific care to clean and improve our translations, resulting in a coverage of \numlangs{} languages with more than 10{,}000 samples per language. We release the dataset to facilitate multilingual RLVR research and benchmarking in the research community.

%% file: content.tex
\def\appendixautorefname{Appendix}

\section{Introduction}
Reinforcement Learning from verifiable rewards (RLVR, \citealp{lambert2025tulu3pushingfrontiers}) has unlocked the latest jump in capabilities of Large Language Models (LLMs), with a particular focus of recent research on the math and logical reasoning domains. In particular, Group Relative Policy Optimization (GRPO, \citealp{shao2024deepseekmath}) coupled with RLVR has been widely applied. For the success of these Reinforcement Learning-based training methods, hitting a ``goldilocks zone'' of difficulty that matches the current policy's capabilities is critical \citep{Polaris2025}. For current state-of-the-art models, which substantially exceed the capabilities of previous generations, this has led to the curation and generation of challenging problem corpora such as AceReason-Math \citep{chen2025acereasonnemotronadvancingmathcode}, DAPO-Math-17k \citep{yu2025dapoopensourcellmreinforcement} or POLARIS-53k \citep{Polaris2025}. 

However, these valuable resources remain limited to English, posing a significant barrier to research of GRPO and RLVR in multilingual settings. 
In this work, we aim to address this gap by providing over 140k high-quality translations of difficult mathematical problems from AceReason-Math \citep{chen2025acereasonnemotronadvancingmathcode}, covering 14 languages with 10k+ samples per language, 7{,}620 of which are parallel across all languages. As human translation at this magnitude carries prohibitive cost, we employ a hybrid approach using both LLM-based translation and human native speaker verification.
This approach allows us to scale our pipeline beyond small-scale datasets. We release the collected data to benefit the research community and facilitate multilingual research in the important new RLVR paradigm.

\begin{table*}[t]
\vspace{-3mm}
\centering
\small
\setlength{\tabcolsep}{6pt}
\begin{tabular}{p{0.47\textwidth} p{0.47\textwidth}}
\toprule
\rowcolor{gray!10}
\textbf{English (Original)} & \textbf{Translated (Language)}  \\
\midrule
How many integers are there between $(11.2)^3$ and $(11.3)^3$?
& \textbf{German:} Wie viele ganze Zahlen gibt es zwischen $(11{,}2)^3$ und $(11{,}3)^3$?\\
\midrule
Given the function $f(x) = x^{2-m}$ is defined on the interval $[-3-m, m^2-m]$ and is an odd function, then $f(m) =$ ?
& \textbf{Italian:} Data la funzione $f(x) = x^{2-m}$ definita nell'intervallo $[-3-m, m^2-m]$ che è una funzione dispari, allora \mbox{$f(m) =$ ?}\\
\midrule
A circle with radius $R$ is inscribed in an isosceles trapezoid. The upper base of the trapezoid is half of its height. Find the area of the trapezoid.
& \textbf{Japanese:} 半径$R$の円が二等辺台形に内接している。この台形の上底は高さの半分である。台形の面積を求めよ。\\
\midrule
Given that an office at a school needs to arrange a duty roster from the 1st to the 6th day with six designated individuals participating, find the total number of different arrangements possible, given that person A and person B cannot be adjacent, and person C and person D also cannot be adjacent.
& \textbf{Chinese:} 学校某办公室需要安排6个指定人员在第1天到第6天进行值班，已知甲和乙不能相邻，丙和丁也不能相邻，求共有多少种不同的安排方案。\\
\midrule
Given a function $f(x)=\log _{a}\left(\sqrt {x^{2}+1}+x\right)+\frac{1}{a^{x}-1}+\frac{3}{2}$, where $a > 0$ and $a \neq 1$. If $f\left(\log _{3}b\right)=5$ for $b > 0$ and $b \neq 1$, find the value of $f\left(\log _{\frac{1}{3}}b\right)$. 
&\textbf{Russian:} Дана функция $f(x)=\log _{a}\left(\sqrt {x^{2}+1}+x\right)+\frac{1}{a^{x}-1}+\frac{3}{2}$, где $a > 0$ и $a \neq 1$. Если $f\left(\log _{3}b\right)=5$ для $b > 0$ и $b \neq 1$, найди значение $f\left(\log _{\frac{1}{3}}b\right)$.\\
\midrule
If 105 customers purchased a total of 300 cans of soda, with every customer purchasing at least one can of soda, determine the maximum possible median number of cans of soda bought per customer that day.
& \textbf{Brazilian Portuguese:} Se 105 clientes compraram um total de 300 latas de refrigerante, sendo que cada cliente comprou pelo menos uma lata de refrigerante, determine a mediana máxima possível do número de latas de refrigerante compradas por cliente naquele dia.\\
\bottomrule
\end{tabular}
\caption{Examples of original samples and their translations to selected target languages from our final dataset.
}
\label{tab:examples}
\vspace{-2mm}
\end{table*}

\section{Related Work}

RLVR \citep{lambert2025tulu3pushingfrontiers} on mathematical reasoning problems has been widely used in the recent literature to study GRPO \citep{shao2024deepseekmath}. 
As an appropriate difficulty-level of the problems to match model capabilities is crucial, researchers have quickly shifted from easier datasets, such as GSM8K \citep{cobbe2021gsm8k}, to datasets with more challenging problems and more available samples \citep{hendrycksmath2021,chen2025acereasonnemotronadvancingmathcode, yu2025dapoopensourcellmreinforcement,Polaris2025}.
This research direction has yielded numerous improvements over the originally proposed GRPO formulation \citep[\emph{e.g.},][]{liu2025understanding, yu2025dapoopensourcellmreinforcement, Polaris2025, deepscaler2025}.
However, these studies have been almost exclusively \textit{English-centric}. Multilingual datasets on the difficulty level of GSM8K exist \citep{shi2022language, chen2023breaking} but efforts to provide more challenging multilingual problem sets have only recently been made \citep{wang2025polymathevaluatingmathematicalreasoning, luo2025mmathmultilingualbenchmarkmathematical}. Still, these more challenging datasets remain limited in size with less than 600 samples per language. This leaves them as valuable resources for multilingual model evaluation but does not enable multilingual GRPO training studies. This gap motivates our work in this paper to create a larger multilingual dataset of challenging mathematical problems.

\section{Dataset Creation}
In this paper, we release a dataset with over 140k high-quality translations of challenging math problems sourced from existing English datasets used for GRPO training \citep{chen2025acereasonnemotronadvancingmathcode}. In the following, we describe the dataset creation process. We show samples from the final dataset in \autoref{tab:examples}.

\subsection{Cleaning the base data}
We manually inspect the original samples from AceReason-Math and observe several patterns of corrupted or unclean data. We distinguish these into salvageable/surface-level issues (we fix and translate these) and critical issues (which exclude samples from further processing). We describe our processing in the following and provide illustrative samples in \autoref{tab:error-categories-detailed}.

\paragraph{Critical issues (\textasciitilde4\%).}\label{sec:data-clean-critical}
We employ a hybrid strategy for filtering out samples, combining programmatic and LLM-based filtering. 
Programmatically, we use custom-written regex expressions to filter out any samples which contain links to images or image placeholders like \texttt{[IMAGE]}. In general, we remove all samples which contain an URL to avoid cases were a question can only be solved with context provided in an external resource. We also search for samples containing further keywords associated with referring to external content -- such as ``Figure'', ``diagram'' -- and filter out any that are problematic in a manual review. We further remove any samples that contain ``\verb|\boxed{[...]}|'', as this was often part of an instruction for the answer format, which might be different from the setup a user of the dataset might want to employ.\footnote{For example, ``\texttt{<answer>[...]</answer>}''.}

For other -- less structured -- failure patterns, we collect examples and inject them into our data cleaning prompt (see \autoref{app:filter-prompt}). These include: (1)~stating the solution in the actual question, rendering the sample useless for RLVR training and benchmarking.\footnote{These solutions could be stripped to keep using the samples but we exclude these for simplicity, as we are not using the entire base dataset for translation in any case.} (2)~Samples that are only solvable with access to referenced supplementary material (\emph{e.g.}, ``Given $f$ as defined previously...''), which is not present. (3)~Questions phrased as ``How can you ...'' when the actual solution given is just a single number (\emph{e.g.}, ``[...] How can we choose $f$ such that $y$ is minimal?'' when the solution given is just the minimal amount). 
(4)~Some samples are in another language than English (most commonly Croatian or Romanian). 

We exclude all samples which are classified as belonging to any of the above categories of critical issues, which amounts to roughly 4\% of our considered original data.

\paragraph{Salvageable issues (\textasciitilde11\%).}
Salvageable issues are more common than critical issues: (1) by far the largest share is made up by ``task annotations'' such as \texttt{Task 5.4:} or \texttt{[21st Math Olympiad]}. These might not be very harmful during RLVR training but we choose to remove these on principle to yield a cleaner dataset. For uniquely identifiable patterns such as \texttt{Task X.X (X points)}, we write regex expression to remove these. However, we do not catch all occurrences, so we also add these to our LLM-based cleaning pipeline. (2) Extraction, \LaTeX{} and formatting issues that are easily correctible based on the context (\emph{e.g.}, \verb|$x 2 + 5 = ?$| when it is clear from context that \verb|$x^2 + 5 = ?$| is correct). (3) Some samples appear to also contain LLM instructions that were used during dataset constructions, \emph{e.g.}, ``Translate the above text into English, please retain the original text's line breaks and format, and output the translation result directly.''. In total, roughly 11\% of the original samples we treat have salvageable issues.

We collect examples for commonly occurring patterns and add these to a ``data-cleaning'' prompt. Before we translate a sample, we first prompt Claude Sonnet 4 to adapt the English version if such salvageable issues are detected. 

Additionally, some samples refer to diagrams provided in the actual question as \verb|[asy]| code. Since solving these tasks is possible but heavily relies on the specialized skill of parsing these \verb|[asy]| instructions, we still translate these samples but provide them in a separate split.

\begin{table*}
\centering
\resizebox{\textwidth}{!}{%
\begin{tabular}{lccccccccccccccc}
\toprule
Model & en & zh & de & es & fr & it & pt & ru & ja & ko & th & bn & te & sw & Average \\
\midrule
Open-weight models (avg@8 evaluation)\\
\midrule

\texttt{gpt-oss-20b} & 79.2 & 78.6 & \textbf{77.5 }& \textbf{75.0} & \textbf{78.3} & \textbf{75.8} & \textbf{76.6 }& \textbf{79.1} & \textbf{78.2} & \textbf{75.5} &\textbf{ 78.2} & \textbf{73.8} & \textbf{77.8} & \textbf{64.8} & \textbf{76.3} $\pm$ 3.4 \\
\midrule

\texttt{SmolLM3-3B} & 77.9 & 53.6 & 69.5 & 72.8 & 72.0 & 70.0 & 71.8 & 69.2 & 58.8 & 55.3 & 59.2 & 28.0 & 16.8 & 20.9 & 56.8 $\pm$ 18.9 \\
\midrule

\texttt{gemma-3-12b-it} & 58.3 & 51.0 & 48.3 & 53.9 & 54.1 & 53.9 & 52.8 & 52.4 & 48.0 & 48.2 & 52.4 & 42.5 & 42.8 & 42.3 & 50.1 $\pm$ 4.6 \\
\texttt{gemma-3-4b-it} & 29.5 & 35.3 & 31.0 & 45.6 & 42.6 & 43.1 & 41.8 & 39.9 & 24.1 & 32.8 & 35.1 & 24.6 & 25.2 & 23.8 & 33.9 $\pm$ 7.2 \\
\texttt{gemma-3-1b-it} & 19.4 & 9.1 & 9.3 & 11.4 & 10.3 & 11.1 & 10.9 & 10.0 & 7.8 & 6.0 & 6.6 & 7.0 & 4.5 & 3.3 & 9.1 $\pm$ 3.6 \\

\midrule

\texttt{DeepSeek-R1-Distill-7B} & {\textbf{83.3}} & {\textbf{80.4}} & {{65.7}} & {{69.1}} & {{70.0}} & {{69.3}} & {{67.7}} & {{68.6}} & {\textbf{62.7}} & {{65.0}} & {{59.6}} & {{67.3}} & {{47.4}} & {{24.2}} & 64.3 $\pm$ 13.3 \\
\texttt{DeepSeek-R1-Distill-1.5B} & 64.8 & 57.5 & 32.6 & 41.4 & 46.2 & 40.3 & 43.8 & 45.0 & 49.5 & 30.8 & 29.4 & 33.6 & 20.8 & 16.6 & 39.5 $\pm$ 12.4 \\

\midrule
Closed models (pass@1 evaluation)\\
\midrule
\texttt{Gemini 2.5 Flash} & {\textbf{78.9}} & {\textbf{85.8}} & {\textbf{81.6}} & {\textbf{79.5}} & {\textbf{82.1}} & {\textbf{81.1}} & {\textbf{81.1}} & {\textbf{82.1}} & {\textbf{81.6}} & {\textbf{84.2}} & {\textbf{80.5}} & {\textbf{80.5}} & {\textbf{83.7}} & {\textbf{76.3}} & \textbf{81.4} $\pm$ 2.2 \\
\texttt{Gemini 2.5 Flash-Lite} & 77.9 & 68.9 & 67.9 & 71.6 & 70.0 & 72.1 & 71.6 & 73.7 & 64.2 & 64.7 & 68.4 & 52.1 & 62.1 & 62.6 & 67.7 $\pm$ 5.9 \\
\midrule
\texttt{Claude Sonnet 4.5} & 76.8 & 78.9 & 75.8 & 78.9 & 76.8 & 76.3 & 73.2 & 80.0 & 74.2 & 77.9 & 76.3 & 71.1 & 71.1 & 72.6 & 75.7 $\pm$ 2.7 \\
\texttt{Claude Haiku 4.5} & 78.4 & 71.6 & 72.1 & 68.9 & 68.4 & 71.1 & 69.5 & 71.1 & 70.0 & 68.9 & 71.6 & 69.5 & 70.0 & 65.3 & 70.5 $\pm$ 2.7 \\
\texttt{Claude Sonnet 3.5} & 52.6 & 50.0 & 48.4 & 48.9 & 50.0 & 51.6 & 51.1 & 54.7 & 47.4 & 51.1 & 48.9 & 38.4 & 40.5 & 36.3 & 47.9 $\pm$ 5.1 \\
\texttt{Claude Haiku 3.5} & 40.0 & 27.4 & 33.2 & 31.1 & 33.2 & 38.9 & 33.2 & 39.5 & 30.5 & 28.9 & 30.5 & 23.2 & 17.4 & 20.5 & 30.5 $\pm$ 6.3 \\
\bottomrule
\end{tabular}
}
\caption{Accuracy (\%) of different models across languages on our dataset. We report pass@1 with a single sampled response per prompt for closed models (classic accuracy) and the average accuracy over eight sampled responses per prompt (avg@8) for open-weight models. The best result for each language within the open- and closed-model categories is marked in \textbf{bold}. The Average column reports mean $\pm$ std across languages.}
\label{tab:multilingual_pass1_accuracy}
\vspace{-4mm}
\end{table*}

\subsection{Translation}
After we have cleaned and filtered the source data, we now enter the translation phase. We provide our prompts in \autoref{app:prompts}.

\paragraph{Initial human translation quality review.} We first create initial translations of a random sample of 100 English problems using Claude Sonnet 4 and conduct a human review with native speaker annotators in German, French, Italian, Spanish, Chinese, Korean, Japanese, Thai, Russian, and Brazilian Portuguese\footnote{Native speaker validation excludes Swahili, Telugu and Bengali. We still provide translations into these languages to increase language coverage.}.
In this round, 97\% of translations were rated as ``acceptable'' (understandable but phrasing could be improved) or ``excellent''. Of the translations that were rated as problematic, a large portion is due to mismatches or changes of \LaTeX{} content compared to the original English version.
We analyze the suggested improvements and issues detected by the native speaker annotators and improve our prompts for the full translation run. 

\paragraph{Full translation run.} For the full translation run, we employ an iterative LLM-based refinement process (also using Claude Sonnet 4). We first create an initial translation as in our initial run (but using the improved prompts). In particular, we prompt to respect number formatting conventions (\emph{e.g.}, US English 12{,}345.67 vs. German 12{.}345,67). Then, the translations are first graded using pre-defined grading criteria similar to our human annotator guidelines and subsequently improved using a refinement prompt,
in case issues are detected. We observed that this approach leads to an improved translation quality and avoids rare but obvious errors in the initial translations. We run this improvement loop up to five times and discard translations which are not accepted in the LLM-based grading round by then. 
The majority of samples do not need to enter this improvement loop and are accepted immediately. We report improvement loop statistics in \autoref{app:additional-stats}.

\subsection{Compiling the final dataset}
\paragraph{Parallel train split.} For research into multilingual model training and evaluation, parallel data across all languages ensures that results are comparable. Therefore, we construct a parallel split which contains only samples for which we have translations that have been accepted in our translation stage for all of our target languages. Our final parallel train split contains 7{,}620 samples per language.

\paragraph{Full train split.} We additionally compile full train splits which contain all available translations for each language, although these splits are no longer parallel across all languages. The full train splits per language contain 10{,}270--12{,}245 samples, varying per language.

\paragraph{Human-validated test set.}
We use a random sample of 190 items\footnote{We start with 200 samples but filtered out samples in an additional posthoc keyword-based cleaning round based on the steps described in \autoref{sec:data-clean-critical}.} which have translations in all our target languages as our test set (these samples are removed from all other splits). We again conduct a human review of these samples with native speaker annotators for the same subset of languages. This time, we render \LaTeX{} content on our annotation platform using MathJax\footnote{\url{https://www.mathjax.org}} to directly display mathematical content instead of \LaTeX{} code, which our annotators were not familiar with.

As in our initial rounds, samples marked as ``problematic'' and suggestions provided by human native speaker annotators in case of ``acceptable'' ratings can be noisy. Issues often stem from lack of mathematical understanding or domain knowledge.\footnote{For example, the following suggested ``improvement'' in German: ``\texttt{sequence}'' should be translated as ``\texttt{Sequenz}'' instead of ``\texttt{Folge}''. Here, ``\texttt{Sequenz}'' would be the literal German translation but ``\texttt{Folge}'' is the appropriate German mathematical term.}
Therefore, we do not directly apply the suggestions but instead filter them selectively with Claude Sonnet 4. We additionally manually check all translations marked as ``problematic''.

We are left with less than 0.1\% of individual translations rated as ``problematic'' (two translations in total, to French and Brazilian Portuguese, respectively).
Additionally, we apply post-editing suggestions by our annotators which were only marked as improving naturalness or stylistic choices rather than critical in 36\% of samples.

\section{Evaluation}

We evaluate several frontier as well as open-weight models on the human-validated parallel test set of 190 samples. We report per-language accuracy in \autoref{tab:multilingual_pass1_accuracy} and provide further details in \autoref{app:eval-details}.  For hosted (closed) models, we sample a single response per prompt and report classic pass@1 accuracy. For open-weight models, we report avg@8, i.e., the average accuracy over eight independently sampled responses per prompt, to reduce variance under stochastic decoding.

Overall, we find consistent trends of larger models exhibiting better performance (\emph{i.e.}, within model families). Additionally, we observe significant differences in performance across languages, particularly in evaluated open-weight models.
We intend to provide a resource to study the effects of RLVR using multilingual training data to complement existing studies which were solely carried out in English (\emph{e.g.}, the original work for which AceReason-Math by \citet{chen2025acereasonnemotronadvancingmathcode} was curated). We provide further evaluation results in \autoref{app:further-eval}.

\section{Conclusion}
We release \textbf{mAceReason-Math}, a multilingual version of a filtered subset of AceReason-Math, covering 14 languages in total and comprising 10k+ translated problems across 13 languages, cleaned versions of the original English problems, a fully parallel train split of 7{,}620 samples per language, and a human-validated test set of 190 samples. Our pipeline combines targeted cleaning with iterative LLM translation and native-speaker checks to produce faithful, high-quality problems.

%% file: content_appendix.tex
\section{Translation Pipeline Prompts}
\label{app:prompts}
We report the prompts used in our LLM-based translation pipeline in the following. We report the prompt used for filtering and cleaning the English base data in \autoref{app:filter-prompt}. In \autoref{app:translation-prompt}, we report the prompt used for initial translations into the target languages. \autoref{app:grading-prompt} contains the prompt used for LLM-based assessment of translations and the generation of improvement suggestions. Finally, \autoref{app:improvement-prompt} contains the prompt used for improving translations based on these detected issues and improvement suggestions.
\subsection{Filtering}
\label{app:filter-prompt}
\begin{minted}[fontsize=\tiny, breaklines=true]{markdown}
# Mathematical Content Quality Assessment & Cleanup

You are analyzing mathematical problem-solution pairs for quality issues and artifacts. This is critical quality control before translation.

## Content to Analyze:
<problem>{problem}</problem>
<solution>{answer}</solution>

## Your Task: 
1. **Assess if severely corrupted** (unsuitable for translation)
2. **Clean up artifacts** (if content is salvageable)  
3. **Assign appropriate tags** based on issues found

## SEVERE CORRUPTION CRITERIA (Filter Out):

### Content Integrity Issues:
- **corruption_solution_given**: Answer provided in the question itself. Examples:
{self._format_examples_section("solution_in_question", 5)}

- **corruption_open_ended**: Questions asking "how can you..." instead of specific answers. Examples:
{self._format_examples_section("open_ended", 5)}

- **corruption_non_english**: Problems not in English. Examples:
{self._format_examples_section("non_english", 1)}

- **corruption_missing_context**: References undefined context or nonsensical Q&A pairs. Examples:
{self._format_examples_section("nonsensical_qa", 1)}

- **corruption_unintelligible**: Completely incoherent problem statements

- **corruption_domain_specific**: Requires ultra-specific non-mathematical knowledge. Examples:
  - Tax problems requiring specific German tax brackets for 2023
  - Problems requiring knowledge of specific company stock prices on exact dates

- **corruption_multiple_unrelated**: Multiple unrelated problems or repeated questions

- **corruption_other**: Severe corruption that doesn't fit any of the above categories. **Use sparingly** and only when content is clearly severely corrupted but doesn't match existing patterns. Must provide detailed corruption_reasons explaining the specific issue.

## ARTIFACT CLEANUP (Fixable Issues):

**CRITICAL SAFETY RULE**: Only remove/fix artifacts that do NOT change the expected solution or important mathematical context. Never alter mathematical meaning.

### Question Artifacts to Clean:
- **cleanup_question_cleaned**: Remove these artifacts from problems:
  - Author attributions: "**Created by [Name]**", "By: [Name]", etc.
  - Creation timestamps or metadata  
  - Platform-specific tags or markers
  - Problem source artifacts like "International Math Olympiad: [Problem]"
  - Graded question formatting like "[Problem] (6 points)" or "## Task 6.2" or "Example 7"
  - Leading task artifacts such as "I-22" in "I-22. Calculate 2+3." or "3-2." in "3.2. Is 234 a prime number?"

- **cleanup_question_fixed**: Fix malformed "question" statements by rephrasing into proper questions. **Use sparingly and only when it significantly improves the problem quality.** Examples:
  - Incomplete statements: "10. The positive integer pairs $(a, b)=$ $\\qquad$ , where $a \\neq b$, that make $\\frac{{a b^{{2}}}}{{a+b}}$ a prime number." → "Find the positive integer pairs $(a, b)$ where $a \\neq b$, that make $\\frac{{a b^{{2}}}}{{a+b}}$ a prime number."
  - Statement format: "The value of $\\sqrt{{16}}$." → "What is the value of $\\sqrt{{16}}$?"
  - Rephrasing unclear statements: "Calculate the area of triangle ABC with vertices A(0,0), B(3,0), C(0,4) using the coordinate formula." → "What is the area of triangle ABC with vertices A(0,0), B(3,0), C(0,4)?"
  
**IMPORTANT**: Only apply cleanup_question_fixed when the original statement is genuinely unclear or confusing. Do not rephrase statements that are already mathematically clear, even if they lack traditional question formatting. When in doubt, leave the problem unchanged.

### Solution Artifacts to Clean:
- **cleanup_solution_cleaned**: Remove these from solutions:
  - Unnecessary LaTeX spacing commands: "\\," (small space), "\\;" (medium space), "\\!" (negative space)
  - Stray formatting characters that don't contribute to mathematical meaning
  - Metadata or attribution text in solutions

- **cleanup_solution_fixed**: Fix obvious extraction errors in solutions (only when correction is completely obvious from context). Examples:
  - Incomplete expressions: "^2=" missing variable → "x^2="
  - Missing variables: "(^2+b^2+^2)" → "(a^2+b^2+c^2)" 
  - Obviously broken notation

**WARNING**: Never "fix" solutions in complex ways requiring actual mathematical reasoning or computation.

## Response Format:

**If severely corrupted:**
```json
{{
  "severity_assessment": {{
    "is_severely_corrupted": true,
    "corruption_reasons": ["Specific reasons"]
  }},
  "tags": ["corruption_solution_given", "corruption_open_ended"],
  "artifacts_found": false
}}
```

**If clean (no issues):**
```json
{{
  "severity_assessment": {{
    "is_severely_corrupted": false,
    "corruption_reasons": []
  }},
  "tags": [],
  "artifacts_found": false
}}
```

**If artifacts cleaned:**
```json
{{
  "artifacts_found": true,
  "changes_made": ["Removed problem numbering 'I-22'", "Fixed incomplete statement to proper question"],
  "cleaned_problem": "Problem after cleanup",
  "cleaned_solution": "Solution after cleanup", 
  "severity_assessment": {{
    "is_severely_corrupted": false,
    "corruption_reasons": []
  }},
  "tags": ["cleanup_question_cleaned", "cleanup_question_fixed"]
}}
```

**Conservative approach**: Only rephrase question / solution when >90%

**Think step by step before giving your final output. Examine every aspect of the content for mathematical coherence, logical consistency, and fundamental integrity before making your assessment.**

Your final output should contain ONLY a valid JSON object within <answer> and </answer> tags
\end{minted}

\subsection{Translation}
\label{app:translation-prompt}
\begin{minted}[fontsize=\tiny, breaklines=true]{markdown}
# Mathematical Problem Translation

You are translating mathematical and logical reasoning problems from English to {language_name}.

## Problem to Translate:
<problem>{task.original_problem}</problem>
<solution>{task.original_answer}</solution>

## Critical Translation Guidelines:

### Mathematical Accuracy (CRITICAL):
1. **Preserve all mathematical concepts**: All variables, numbers, equations, and logical relationships must remain identical
2. **Maintain problem structure**: The mathematical logic and problem logic must be unchanged
3. **Verify mathematical consistency**: Ensure translated problem leads to the same solution method

### LaTeX Handling (CRITICAL):
1. **NEVER translate LaTeX content**: Preserve all content within `$...$` or `\\(...\\)` exactly as written
2. **Preserve exact spacing**: Match LaTeX spacing and notation style precisely (e.g., do not change `$\\triangle A B C$` to `$\\triangle ABC$`)
3. **Within LaTeX, only translate \\text{{}} content**: Within LaTeX, only translate natural language inside `\\text{{...}}`
4. **Keep LaTeX environments**: Maintain `$...$`, `\\(...\\)`, `\\begin{{equation}}...\\end{{equation}}` notation exactly and do not switch in the translation
5. **DO NOT TRANSLATE mathematical or other language-agnostic content**:
    - Pure numbers, mathematical expressions, formulas, symbols or notation: "42", "-7", "0", "x²", "sin(π/4)", "y = mx + b", ...
    - Keep LaTeX or mathematical markup such as: "\\frac{{}}{{}}", "\\angle{{}}", "\\overline{{}}", "\\triangle{{}}", ...

### Natural Language Quality (CRITICAL):
1. **Create natural flow**: Use natural {language_name} sentence structure, not word-for-word translation. A natural translation that restructures sentences flow or word order is better than an awkward literal word-for-word translation.
2. **Use informal register**: Use informal forms (e.g. German: "Du" instead of "Sie", French: "tu" instead of "vous", etc.)
3. **Apply number formatting**: Use {language_name} number formatting conventions (e.g. for German use 12.345,60 instead of the English: 12,345.60)
4. **Standard mathematical terminology**: Use terms familiar to native {language_name} speakers. For jargon or technical terms, use the terms that would be commonly used in {language_name} higher education contexts.
5. **Cultural Context**: If it does not influence the logic of the problem, localize English-centric references to a {language_name} cultural context. 

### Answer Format Requirements:
1. **Always use Arabic numerals**: Answers must use 0-9 digits regardless of target language
2. **Apply number formatting**: Use {language_name} number formatting conventions (e.g. for German use 12.345,60 instead of the English: 12,345.60)
3. **Translate natural language**: Convert "True"/"False"/"Increasing" etc. to {language_name}
4. **Keep mathematical notation unchanged**: Preserve LaTeX, formulas, expressions exactly
5. **Match original LaTeX style**: Don't add or remove LaTeX formatting unless number conventions differ

### Language-Specific Guidelines for {language_name}:
{self._get_language_specific_guidelines(task.language)}

## Response Format:
Respond with a JSON object containing:
- `translated_problem`: string (the problem translated to {language_name})
- `translated_solution`: string (translated if natural language, unchanged if mathematical)

First, think step by step. Then output your final JSON translation within <answer> and </answer> tags.
\end{minted}

\subsection{Grading}
\label{app:grading-prompt}
\begin{minted}[fontsize=\tiny, breaklines=true]{markdown}
# Mathematical Problem Translation Quality Assessment

You are assessing the quality of a mathematical problem translation from English to {language_name}.

## Original (English):
<problem>{task.original_problem}</problem>
<solution>{task.original_answer}</solution>

## Translation ({language_name}):
<problem>{translated_problem}</problem>
<solution>{translated_solution}</solution>

## Assessment Criteria (Based on Annotation Guidelines):

### Critical Issues (Must be perfect):
1. **Mathematical Accuracy**: All mathematical concepts, relationships, and logical structures preserved exactly
2. **LaTeX Consistency**: 
   - All `$...$` and `\\(...\\)` content identical to original
   - Spacing within LaTeX matches exactly (e.g., `$\\triangle A B C$` spacing preserved)
   - Only `\\text{{...}}` content translated within LaTeX
3. **Logical Consistency**: Translation maintains same mathematical relationships as original. No new details should be "invented"
4. **Answer Format**: 
   - Natural language properly translated to {language_name}
   - Mathematical notation unchanged from original
   - Arabic numerals used correctly
   - Answer matches question if cultural or other details in the question were changed to be more appropriate for {language_name}

### Critical Quality Considerations:
1. **Mathematical Terminology**: Uses appropriate {language_name} mathematical terms commonly used in {language_name} higher education
2. **Natural Fluency**: Sounds natural to native {language_name} speakers (no "translationese"). A natural translation that restructures sentences flow or word order is better than an awkward literal word-for-word translation
3. **Register Appropriateness**: Uses informal register correctly 
4. **Number Formatting**: Applies {language_name} conventions appropriately

### Important Checks:
- **Terminology Accuracy**: Flag non-standard or incorrect mathematical terms
- **LaTeX Preservation**: Verify exact match with original LaTeX formatting
- **Natural Flow**: Assess if translation sounds natural in {language_name}, avoid overly literal and stilted translations
- **Cultural Appropriateness**: Check mathematical conventions for {language_name}

## Required Response Format:
Respond with a JSON object containing:
- `translation_quality_is_excellent`: boolean (true only if NO issues found)
- `concrete_issues`: array of strings (specific issues with format: "Issue: [description]. Suggestion: [fix]")
- `mathematical_accuracy_preserved`: boolean (true if all math concepts preserved)
- `answer_handling_correct`: boolean (true if answer was handled correctly)

For terminology issues, use format: "Mathematical terminology: '[English term]' should be translated as '[correct {language_name} term]' instead of '[current translation]'"

First, think step by step. Then, as your final answer, output ONLY a valid JSON object within <answer> and </answer> tags.
\end{minted}

\subsection{Translation improvement}
\label{app:improvement-prompt}
\begin{minted}[fontsize=\tiny, breaklines=true]{markdown}
# Mathematical Problem Translation Improvement

You are improving a mathematical problem translation from English to {language_name} based on specific identified issues.

## Original (English):
<problem>{task.original_problem}</problem>
<solution>{task.original_answer}</solution>

## Current Translation ({language_name}):
<problem>{current_translation.get("translated_problem", "")}</problem>
<solution>{current_translation.get("translated_solution", "")}</solution>

## Improvement Guidelines (Based on Annotation Guidelines):

### Priority 1 - Critical Fixes:
1. **Mathematical Accuracy**: Ensure all mathematical concepts preserved exactly
2. **LaTeX Consistency**: Fix any spacing or notation inconsistencies with original
3. **Logical Preservation**: Maintain identical mathematical relationships

### Priority 2 - Quality Enhancements:  
1. **Natural Fluency**: Create natural {language_name} while preserving mathematical meaning
2. **Mathematical Terminology**: Use standard {language_name} mathematical terms used in higher education contexts
3. **Register Correction**: Apply informal forms correctly
4. **Number Formatting**: Use {language_name} conventions appropriately

### LaTeX Correction Rules:
- Match original spacing exactly: If English has `$\\triangle A B C$`, the translation should also use `$\\triangle A B C$` instead of e.g. `$\\triangle ABC$` 
- Preserve notation style: `\\(...\\)` vs `$...$` must match original
- Only translate `\\text{{}}` content within LaTeX
- Never alter mathematical expressions or formulas

### Language-Specific Guidelines for {language_name}:
{self._get_language_specific_guidelines(task.language)}

## Issues to Fix:
{issues_text}

## Required Response Format:
Respond with a JSON object containing:
- `improvement_notes`: string (explanation of changes made)
- `translated_problem`: string (improved problem translation)
- `translated_solution`: string (improved solution - translated if natural language, unchanged if mathematical)

First, think step by step. Then output ONLY the improved JSON translation within <answer> and </answer> tags.
\end{minted}

\section{Additional Statistics}
\label{app:additional-stats}
\begin{figure*}
    \centering
    \includegraphics[width=.8\linewidth]{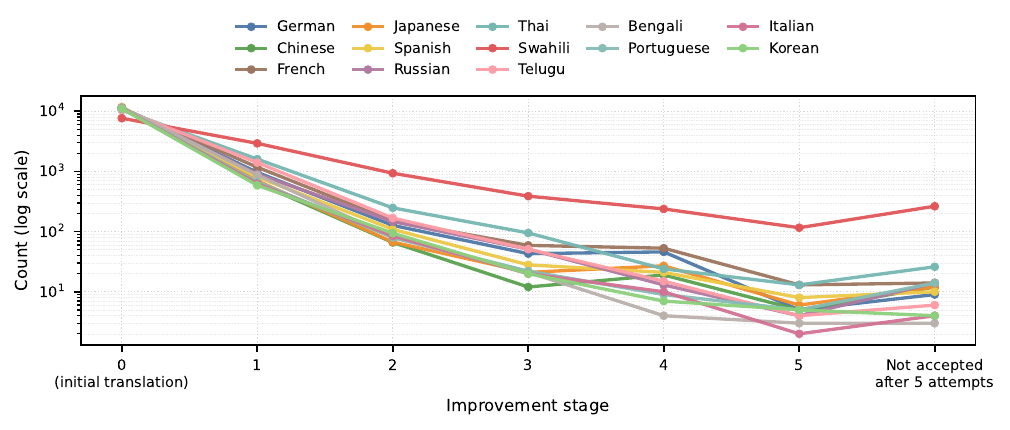}
    \caption{Number of accepted samples per improvement stage in the translation pipeline. Most samples are accepted after the initial translation or one additional round of LLM-based refinement.}
    \label{fig:improvement_histogram}
\end{figure*}

We report the exact number of samples per split and language in mAceReason-Math in \autoref{tab:dataset_stats}. We further report the number of accepted samples per improvement stage in our translation pipeline in \autoref{fig:improvement_histogram}.

\begin{table}
\centering
\resizebox{0.5\linewidth}{!}{%
\begin{tabular}{lccc}
\toprule
Language & Train (Parallel) & Test (Parallel) & Train (All) \\
\midrule
English     & 7,620 & 190 & 12,245 \\
German      & 7,620 & 190 & 11,151 \\
French      & 7,620 & 190 & 11,007 \\
Spanish     & 7,620 & 190 & 11,346 \\
Chinese     & 7,620 & 190 & 10,470 \\
Russian     & 7,620 & 190 & 11,237 \\
Japanese    & 7,620 & 190 & 10,376 \\
Swahili     & 7,620 & 190 & 11,124 \\
Telugu      & 7,620 & 190 & 10,964 \\
Bengali     & 7,620 & 190 & 11,082 \\
Thai        & 7,620 & 190 & 11,104 \\
Portuguese  & 7,620 & 190 & 10,632 \\
Italian     & 7,620 & 190 & 10,646 \\
Korean      & 7,620 & 190 & 10,270 \\
\bottomrule
\end{tabular}
}
\caption{Number of samples in mAceReason-Math per split and language.}
\label{tab:dataset_stats}
\end{table}

\section{Data Cleaning Examples}
We provide an overview of our filtering criteria for excluding samples present in the original AceReason-Math dataset \citep{chen2025acereasonnemotronadvancingmathcode} in \autoref{tab:error-categories-detailed}.

  \begin{table*}[htbp]
    \centering
    \caption{Overview of ``critical issues'' identified in the original data. Examples abbreviated with [\ldots] for space constraints. \textbf{Bold text} highlights key corruption indicators
  within the quoted examples. }
    \label{tab:error-categories-detailed}
    \renewcommand{\arraystretch}{1.4}
    \begin{tabularx}{\textwidth}{@{}p{3cm}@{\hspace{0.6cm}}c@{\hspace{0.6cm}}p{2cm}@{\hspace{0.4cm}}X@{}}
    \toprule
    \textbf{Primary Category} & \textbf{Count} & \textbf{Subcategory} & \textbf{Illustrative Examples} \\
    \midrule
    \multirow{9}{3cm}[9ex]{\textbf{Missing Context}} & \multirow{9}{*}[9ex]{\textbf{\textasciitilde2\%}} & \textit{Missing\newline{}Visual} &
    ``The diagram shows three rectangles and three straight lines. What is the value of p + q + r [...]'' \newline \footnotesize{(No diagram provided)} \\
    \cmidrule(lr){3-4}
    & & \textit{Missing\newline{}Definitions} & ``Define a regular n-pointed star as described in the original problem [...]'' \newline \footnotesize{(Critical definitions not provided)} \\
    \midrule
    \multirow{12}{3cm}[11ex]{\textbf{Content and\\ Extraction Issues}} & \multirow{12}{*}[11ex]{\textbf{\textasciitilde1.5\%}} &
  \textit{Answer\newline{}Leakage} &
    ``[...]\textbf{ Answer:} decreased by 6\%'' \newline \footnotesize{(Solution provided in problem)} \\
    \cmidrule(lr){3-4}
    & & \textit{Processing\newline{}Artifacts} & ``[...] text above has been \textbf{translated as requested} [...]'' \newline
    \footnotesize{(Leftover data-processing artifacts)} \\
    \cmidrule(lr){3-4}
    & & \textit{Truncated\newline{}Content} & ``[...] We know that for any pair of distinct numbers from this set there is'' with answer: ``50'' \newline \footnotesize{(Cuts off before finishing question, answer is just ``50'')} \\
    \midrule
    \multirow{6}{3cm}[4ex]{\textbf{Ill-posed Question}} & \multirow{6}{*}[4ex]{\textbf{\textasciitilde0.5\%}} & \textit{Free-text\newline{}answer} &
    ``\textbf{How can you} cut a 5×5 square [...] into 50 equal squares?'' \newline \footnotesize{(No methodology
    provided, answer is just a single integer)} \\
    \cmidrule(lr){3-4}
    & & \textit{Mathematical\newline{}Errors} & ``$x+\frac{1}{x}$ has a \textbf{maximum}'' \newline \footnotesize{(Impossible
    claim)} \\
    \midrule
    \multirow{2}{3cm}[1ex]{\textbf{Other Issues}} & \multirow{2}{*}[1ex]{\textbf{<0.01\%}} & \textit{Various} &
  ``Odredi sve prirodne brojeve $n$ za koje postoje [\ldots]'' \newline \footnotesize{(\emph{e.g.}, non-English problems, ultra-specific domain knowledge, ...)} \\
    \bottomrule
    \end{tabularx}
    \end{table*}

\section{Evaluation}
\subsection{Evaluation Details}
\label{app:eval-details}
For evaluating hosted models, we sample a single response per prompt (pass@1). We conduct the evaluations in October and November 2025. 

To extract answers, we prompt the models to provide the final answer in \texttt{<answer>...</answer>} tags. This instruction is appended at the end of the prompt and given in the prompt's language (translated from English). We first attempt to extract an answer from the \texttt{<answer>...</answer>} tags but additionally attempt to extract an answer from \verb|\boxed{...}| if no \texttt{<answer>...</answer>} tags are present. We then evaluate the extracted answer against the original ground-truth using \href{https://github.com/huggingface/Math-Verify}{\texttt{math-verify}}. Additionally, since in some cases the answer has been adapted in the translated version, \emph{e.g.}, to match the German number formatting conventions, we also check for string equality of the extracted answer with the translated version's answer.

\subsection{Further Evaluation Results}
\label{app:further-eval}
We provide further evaluation results for the \texttt{Qwen3} model family in \autoref{tab:results_avg_appendix}. The \texttt{Qwen3} reasoning models (``non\texttt{-Base}'') in particular achieve strong performance across many languages, even outperforming frontier models such as \texttt{Claude Sonnet 4.5}. This suggests that our data might have been in the training mix of the \texttt{Qwen3} reasoning models during RLVR training. Furthermore, since the \texttt{Qwen3} models achieve good performance also in non-English languages, it suggests that math-problem solving capability exhibits strong cross-lingual transferability at the problem instance level. We investigate this in future work.
\input{results_appendix}

%% file: results_appendix.tex
\begin{table*}
\centering
\resizebox{\textwidth}{!}{%
\begin{tabular}{lccccccccccccccc}
\toprule
Model & en & zh & de & es & fr & it & pt & ru & ja & ko & th & bn & te & sw & Average \\
\midrule
Open-weight models (avg@8 evaluation)\\
\midrule

\texttt{Qwen3-14B-Base} & 31.4 & 29.5 & 24.9 & 27.3 & 25.2 & 21.7 & 25.9 & 28.2 & 21.2 & 20.2 & 18.8 & 13.2 & 12.1 & 5.7 & 21.8 $\pm$ 6.8 \\
\texttt{Qwen3-8B-Base} & 25.8 & 23.2 & 17.8 & 21.8 & 20.7 & 18.6 & 22.9 & 23.9 & 15.3 & 14.4 & 12.5 & 9.0 & 12.6 & 2.8 & 17.2 $\pm$ 6.1 \\
\texttt{Qwen3-4B-Base} & 15.5 & 15.3 & 12.1 & 13.6 & 13.6 & 13.6 & 14.5 & 13.3 & 10.3 & 8.3 & 7.9 & 8.1 & 8.4 & 3.0 & 11.2 $\pm$ 3.4 \\
\texttt{Qwen3-1.7B-Base} & 8.0 & 5.0 & 3.0 & 3.8 & 4.0 & 3.2 & 4.2 & 4.1 & 2.3 & 2.6 & 2.4 & 0.9 & 0.9 & 0.5 & 3.2 $\pm$ 1.8 \\
\midrule
\texttt{Qwen3-14B} & {\textbf{88.8}} & {\textbf{89.9}} & {\textbf{91.6}} & {\textbf{91.3}} & {\textbf{90.9}} & {\textbf{92.0}} & {\textbf{91.1}} & {\textbf{87.5}} & {\textbf{89.3}} & {\textbf{89.2}} & {\textbf{90.5}} & {\textbf{90.6}} & {\textbf{87.6}} & {\textbf{69.7}} & \textbf{88.6} $\pm$ 5.2 \\
\texttt{Qwen3-8B} & 85.4 & 84.9 & 86.4 & 87.9 & 87.3 & 87.4 & 85.9 & 85.5 & 86.1 & 85.1 & 85.7 & 86.0 & 82.8 & 57.8 & 83.9 $\pm$ 7.1 \\
\texttt{Qwen3-4B} & 85.0 & 85.6 & 88.6 & 87.2 & 87.4 & 86.9 & 88.0 & 82.8 & 83.6 & 83.7 & 84.7 & 84.1 & 78.1 & 44.2 & 82.1 $\pm$ 10.5 \\
\texttt{Qwen3-1.7B} & 73.1 & 71.9 & 73.4 & 74.3 & 75.7 & 75.7 & 74.1 & 71.1 & 72.1 & 69.9 & 69.3 & 64.8 & 57.8 & 26.1 & 67.8 $\pm$ 12.0 \\

\bottomrule
\end{tabular}
}
\caption{Accuracy (\%) of different models across languages on our dataset. We report pass@1 with a single sampled response per prompt for closed models (classic accuracy) and the average accuracy over eight sampled responses per prompt (avg@8) for open-weight models. Best result per language is marked in \textbf{bold}. The Average column reports mean $\pm$ std across languages.}
\label{tab:results_avg_appendix}
\vspace{-4mm}
\end{table*}